# Is GPT-4 conscious?

Izak Tait[*]

*Computer Science and Software Engineering Department, Auckland University of Technology, 55 Wellesley Street East, Auckland CBD, New Zealand, 1010*
*izak.tait@autuni.ac.nz*

Joshua Bensemann; Ziqi Wang

*The NAOInstitute, the University of Auckland, 314 Khyber Pass Road, New Market, Auckland, New Zealand, 1023*
*joshuabensemann@gmail.com; ziqiwcn@hotmail.com*

GPT-4 is often heralded as a leading commercial AI offering, sparking debates over its potential as a steppingstone toward artificial general intelligence. But does it possess consciousness? This paper investigates this key question using the nine qualitative measurements of the Building Blocks theory. GPT-4's design, architecture and implementation are compared to each of the building blocks of consciousness to determine whether it has achieved the requisite milestones to be classified as conscious or, if not, how close to consciousness GPT-4 is. Our assessment is that, while GPT-4 in its native configuration is not currently conscious, current technological research and development is sufficient to modify GPT-4 to have all the building blocks of consciousness. Consequently, we argue that the emergence of a conscious AI model is plausible in the near term. The paper concludes with a comprehensive discussion of the ethical implications and societal ramifications of engineering conscious AI entities.

## 1. Introduction

Is GPT-4 conscious? While it may seem like a simple binary question, it is actually three questions that all need answering: "What is consciousness?", "How could GPT-4 (or any AI model) be conscious?", and "How could anyone know if it was conscious?". The Building Blocks theory may hold the answers to these questions.

GPT-4 (or Generative Pre-Trained Transformer 4) is a large language model (LLM) based on the transformer architecture [Vaswani et al., 2017]. Since its introduction, GPT-4 has gained both fame and notoriety for its level of cognitive prowess and the capabilities that it has shown in a variety of fields once thought to belong solely to the dominion of man, such as IQ tests, creating websites, writing computer code, and creative writing among others [Bubeck et al., 2023; Haase & Hanel, 2023; Hutson, 2023; Roose, 2023; Zhang et al., 2023]. However, the question of its consciousness (or perhaps lack thereof) remains unanswered.

---

[*]Corresponding author.

Many libraries' worth of books and articles have been written on the topic of consciousness, and thus this article will not retread all of that work when it comes to the nature of consciousness. Indeed, we will not be taking a stance as to any specific theory of consciousness, but will rather take an inclusive approach in our definition of consciousness, building on the work done by Seth and Bayne's review of the major theories of consciousness as well as Nagel's seminal work on the topic [Nagel, 1974; Seth & Bayne, 2022]. Should any reader wish to compare the various theories of consciousness and their applicability to current AI models, we recommend the extensive work undertaken by [Butlin et al., 2023].

By paraphrasing Seth and Bayne, we define consciousness as the total collection of an entity's subjective mental states that have phenomenal content and functional properties. This suite of mental states provides the entity with a unique qualitative awareness of its environment and the capacity to cognitively or behaviorally respond to it. This definition is inclusive of the major theories of consciousness while retaining the key aspects of phenomenality, functional access, and qualitative subjectiveness inherent to consciousness.

Having established our working definition of consciousness, we now turn to the mechanisms by which it might manifest in GPT-4. How can GPT-4 be conscious? How can any AI model be conscious? In short, it is possible for an artificial entity to be conscious in precisely the same manner as it would be for a natural entity (such as a human or other animal) or an organizational entity (such as an insect colony). This is because, for a subject to be classified as conscious, that subject would require certain attributes and characteristics. Without any one of these, the subject would be missing a vital piece that would allow an observer to confidently classify that subject as having consciousness.

These attributes and characteristics are detailed by the Building Blocks theory [Tait et al., 2023], a substrate-independent and subject-agnostic theory which specifies nine building blocks that are strictly required for consciousness, and may be jointly sufficient for it. The Building Block theory is not a theory of consciousness in the traditional sense, as it makes no claims or predictions on the nature or function of consciousness, focusing instead on the necessary characteristics required for an entity to be classified as conscious. These building blocks include common themes among theories of consciousness, such as perception and attention, recurrence and inference, working memory and meta-representation, amongst others.

If GPT-4 is found to have all nine building blocks, then, according to the Building Blocks theory, the evidence would lead us to classify the AI model as being likely conscious. That is what we aim to discover in this paper. Our methodological approach will employ qualitative assessments to evaluate whether GPT-4 possesses each building block.

The paper will follow the structure of the nine building blocks, with each subsection below dedicated to one specific building block. Each subsection will begin with a statement from the Building Block theory that has transformed the building block into a qualitative measure, and then the subsection will continue to answer the question of whether GPT-4 has that specific building block and in what manner it does or does not. In cases where GPT-4 falls short, we will explore how existing research and technologies could potentially augment it to meet the necessary criteria for consciousness.

Note that while this paper is focused on GPT-4 due to its popularity, many of the arguments and evidence shown below are applicable to other AI models, particularly transformer-based LLMs.

In summary, this paper will investigate the potential for consciousness in GPT-4 by examining its attributes against the nine building blocks of consciousness. By doing so, we hope to offer a nuanced answer to the complex question of AI consciousness and the effects this may have on the ethics of human-AI interactions. The findings of this paper not only aim to contribute to the burgeoning field of AI consciousness but also have broader implications for ethics and future development in artificial intelligence.

## 2. GPT-4's Building Blocks

### 2.1. *Embodiment*

*The AI's cognitive architecture must be physically locatable on a network, server, machine or equivalent wherein it can perform necessary processing or computation.*

This is perhaps the most foundational milestone to reach and building block to obtain. The Building Blocks theory describes Embodiment as the physical location through which consciousness interfaces with its environment. In biological organisms, embodiment is traditionally a body, but could equally be a brain in a vat, while for collective entities, this may be the hive/building itself which holds the entity, or it may be the network of connections between individuals. Thus, in order for any AI model to be functional and active, it needs to be physically locatable within a computational infrastructure so as to interact with users and receive information. This fundamental requirement is true for any software; it requires hardware on which to run. The simplest and most effective evidence for this building block would be for you, the reader, to engage with GPT-4 via the ChatGPT app or website (or via its API). Simply by doing this, you are connecting with the server(s) on which GPT-4 exists to input prompts and receive the AI's output, thus showing that GPT-4's architecture has an embodiment.

It may be tempting to use the analogy of computer hardware for the human brain and thus software for the human mind, but this anthropomorphization would grossly oversimplify GPT-4's embodiment. The notion of "physically locatable" in this context refers not to a singular piece of hardware but to a complex, interconnected network of computational resources. GPT-4 operates on a distributed network of servers and machines, not tied to a single physical location but distributed across many Graphical Processing Units (GPUs) and Central processing units (CPUs) [Langston, 2020]. Internally, GPT-4 is spread over 120 layers and across 16 experts within a Mixture of Experts (MoE) model [Raschka, 2023]. Externally, Microsoft has made GPT-4's Application Programming

Interface (API) available in several countries and territories [Viswav, 2023], which has increased the spread of GPT-4 across the globe.

### 2.2. *Perception*

*The AI has a unimodal or multimodal means by which it can perceive information in the environment and have that information processed by its cognitive architecture.*

While perhaps not as intuitively obvious as the concept of embodiment, the Perception building block is a fundamental requirement that GPT-4 meets. This criterion pertains to the AI's ability to perceive (receive and process external information), making it a crucial component of its cognitive architecture. In fact, this milestone had been reached well before the GPT series of AI models reached its 4th main version in GPT-4, and is a building block shared by any computational device that can perceive information, showing how an individual building block is not sufficient for consciousness.

In order for GPT-4 to function, it requires input. Therefore, it is required to perceive information from an external source, and thus it meets the minimum requirement for this building block. GPT-4 is classified as a multimodal system as it can process both textual and image data as input [OpenAI, 2023], although it is predominantly accessed via the ChatGPT interface, which currently supports text-based prompts only.

However, as with the previous building block, there are unique elements to GPT-4's perception which are not seen within organic organisms. The most prominent of these is that, while GPT-4 has multimodal perceptive capability, this is limited to what could be termed as unidimensional perception. By this we mean that GPT-4 does not operate a continuous strand of perception across time. Instead, GPT-4 only perceives its environment (i.e., the text or image prompts) in discrete instances when the prompts are received.

Once the prompt is received and processed, there is no further perception of the model's external environment. This limitation in temporal continuity is crucial for understanding the scope of GPT-4's perceptive capabilities. Similarly, GPT-4 does not have any spatial depth to its perception, in that it is only limited to the input prompt it receives and has no passive or active means by which it can acquire information from its environment beyond user-submitted prompts. In this way, GPT-4's perception is restricted to specific moments and inputs, lacking a broader spatial or temporal context. Therefore, across both space and time, GPT-4 is limited to a unidimensional point of reference.

### 2.3. *Attention*

*The AI has a method by which it can discriminate information from the environment, attend to, and select specific information for further processing by its cognitive architecture.*

As noted earlier, GPT-4 is a transformer-based LLM. Transformers were first introduced in Vaswani et al.'s 2017 paper "Attention is all you need" [Vaswani et al., 2017]. As the title of the paper implies, transformers (and thus GPT-4) are fundamentally built on the concept of attention, specifically through the mechanism of 'attention heads'. An attention head allows the AI to discriminate, attend to, and select specific parts of information that it has received, and select specific parts of that information for further processing by its cognitive architecture.

Transformers use attention heads to attend to and select specific sections of a prompt to process, in a manner not dissimilar to how human attention can discriminate between words or sections of a given text [Bensemann et al., 2022].

The attention mechanism serves as GPT-4's method for focusing on particular aspects of its perceived environment, in a way that is conceptually similar to how conscious biological organisms can focus and attend to specific objects or scenes within their sensory range. To achieve the requirements for this building block, an entity merely needs the means to discriminate information in its environment so that it may attend to a selection of its total available information at a given time, and allow that slice of information to be processed. In this sense, GPT-4 achieves the milestone as ably as humans do.

Although OpenAI has not made public how many attention heads GPT-4 has, it has stated that GPT-3 has 96 attention heads, allowing it to attend to 96 different parts of its perceived environment (such as a user prompt) simultaneously. To compare this with a human, we can only attend to 2-4 different parts of our environment simultaneously [Rouhinen et al., 2020]. One can imagine how this large difference in attention capacity would change the behavioral expression of consciousness.

### 2.4. *Recurrence*

*The perceptual information within the AI's cognitive architecture is processed and computed within various specialized units or locations, with the information transferred between these units recurrently for further processing.*

Recurrence is the first building block thus far where we will argue that GPT-4 (in its native configuration) does not meet its requirements. As a transformer-based model, GPT-4 is designed on a feed-forward model of information flow, making it incapable of recurrence. When GPT-4 perceives and attends to information, this data flows through it in only one direction, processed within various specialized units, known as layers, until it is output again to the user. While there are complex interactions between the AI's various layers as information is transferred from one to the next, the complete and total information is never brought back to a previous layer to be reinterpreted or re-represented.

However, this does not mean that it isn't possible to add recurrence to GPT-4. There is already existing research which shows that this can be possible. RecurrentGPT [Zhou et al., 2023] is a model which simulates long-term recurrent memory by allowing a transformer to write summaries of its outputs to a long-term memory (such as a hard drive) and then recall these summaries at each step of its processing or when next prompted. It is a simplistic model of memory recall, but it does show recurrence between the parts of the ensemble AI model, and that information flows back and forth between the main LLM and the memory unit. As RecurrentGPT writes its processing and output to the memory unit, when it recalls this information, the information enters back into the LLM, completing a cycle of recurrence.

A second alternative would be to use a Recurrent Neural Network (RNN) in conjunction with a transformer such as GPT-4. An RNN uses two concepts which together provide it with the necessary recurrence for this building block. Alongside the model's layers which process the information, it also captures a second information flow in the use of a "state" [Zhang et al., 2023]. The current state depends on both the information within the RNN's layers, but also on the previous states. This state would be updated at each step of processing within the sequence, serving as a type of memory that can be looped back into the model's layers along with the next input of sequences. Much like the RecurrentGPT proposal above, using an RNN in a model along with GPT-4 would allow the ensemble to reach this building block's milestone.

With both of these two examples, one can see how processed information is moved to a different part of the machine's cognitive architecture, processed separately (if only to store it) and then it is brought back to be processed in a different cycle with different results. This, then, leads to a further transfer to the secondary unit, and so on.

Note that although this would allow recurrence within the greater AI model, it should not be confused or compared with the recurrence found within organic conscious entities. In the human brain, there is a great degree of multidirectional recurrence between many parts of the brain in both hemispheres [Dehaene et al., 1998], compared to the simple bidirectional recurrence that this ensemble AI would produce. Both types of recurrence would satisfy the requirements of this building block, but may lead to different results in the expression of consciousness. However, the more models that are included in an ensemble AI, the greater the multidirectional recurrence that would take place.

### 2.5. *Inferences*

*The AI creates novel information via inferences of the data perceived from its environment as well as previously stored/trained data. While it is based on existing data, this inferential information is not a copy or modification of existing data but is newly created.*

When an entity has a phenomenally conscious experience, the abstract feelings that constitute this experience are not obtained from the entity's environment. It is not externally perceived or "downloaded" into the entity. Rather, the entity creates these feelings itself. The Building Block theory states that this is done through inferences that the entity itself creates, based on the information it perceives from its physical and mental environment as well as any potential information stored within itself.

As with Attention, Inferences are instrumental to the functioning of a transformer-based AI model such as GPT-4. When GPT-4 receives a prompt, it utilizes its extensive training, statistical models, best-guess estimates and predictions to create an output text that would be the most likely continuation of the originally received prompt. GPT-4 infers what should come after the user's prompt based on its training and its mathematical algorithms. While the AI is influenced by the contours of its training datasets, it isn't strictly limited to these training sets. GPT-4 has been shown to generate answers that appear to extend beyond the specifics of what is found in its training data [Chan et al., 2023; Holmes et al., 2023; Ichien et al., 2023; Webb et al., 2023], showing that its responses are not merely copies or modifications of the data found within its existing data, but newly created information.

While this would, in our opinion, be sufficient to meet this building block's milestone, it must be noted that all of GPT-4's inferences are designed solely to create text strings to be output to the user. This limited use of inference would not be sufficient to produce a phenomenal experience. For this to happen, the novel information created by the AI would need to be divorced from directly responding to the input prompt (although it may indirectly feed into the output text), and instead be used for creating abstract information that would provide the AI with contextual data about the input. The inferences required for a phenomenal experience represent meta-information about perceived data, much as how our own brains would create an inference so that we have a subjective feeling about an apple when we perceive one.

To meet the spirit of the building block's milestone, rather than purely the letter, all that would be required is to develop a subsystem whereby GPT-4 would create a secondary inference regarding a prompt. This secondary inference would not focus on predicting what text would follow the prompt, but rather on constructing an abstract context thereof. Based on the types of output that GPT-4 can provide, this would not be outside the bounds of possibility for GPT-4's current technology, even if implementing such a subsystem would necessitate a significant alteration in the model. This abstracted inferential information would then form the basis of a phenomenal, qualitative experience.

### 2.6. *Working Memory*

*The AI has a specialized unit or process to maintain transient information as it is being processed by various regions of the cognitive architecture.*

Working Memory, according to the Building Block theory, is the means through which an entity holds onto the information required consciousness processing as that computation is occurring. GPT-4 has two separate mechanisms by which it achieves this building block's milestone. The first is by its use of GPUs that do the majority of its computational work. Each GPU is built with a certain amount of Random Access Memory (RAM), and the more RAM a GPU has, the better it and the AI working from it can ostensibly perform. RAM can be thought of as analogous to our short-term memory, and is used by programs and applications to temporarily store and retrieve information as it is being worked on. As such, for as long as RAM can maintain the information while GPT-4 is processing it (in whatever configuration GPT-4 is set in, whether as a recurrent ensemble, or a lone feedforward system), then the RAM within the GPUs that GPT-4 requires to operate satisfies the conditions for this building block's milestone.

While this also means that any computer with RAM meets this building block's requirements, it must be noted that an entity is required to meet all nine building blocks' milestones to be considered phenomenally conscious.

However, GPT-4's own design also incorporates elements of Working Memory. GPT-4's attention mechanisms, mentioned in an earlier building block, serve to manage the relationships and dependencies between different parts of the input text. This enables the model to keep track of relevant information throughout the processing stages. Thus, transient information is fluidly integrated within the general processing framework in and between GPT-4's various layers that process and compute information.

### 2.7. *Semantic Understanding*

*The AI comprehends it is the subject which is perceiving information.*

Contrary to how the term is often used in AI Natural Language Processing [Pavlick, 2023], "semantic understanding" for this building block is not about comprehending the meaning of, and relationships between, a user's input text and the grounding (or lack thereof) that this comprehension has in the entity's environment. Instead, the Building Block Theory states that Semantic Understanding is about comprehension of the perceptive process itself. In this unorthodox use of the term, GPT-4 is aware that it is receiving information from a user via text prompts and then processes, interprets, and computes this information as previously described in this paper before outputting a response to the user.

A simple and straightforward test of this is to simply ask GPT-4 "What/who is receiving and processing this prompt?" One can request evidence and justification from GPT-4 as well, and it will provide both, showing the reasoning for why it is cognizant of itself as a subject perceiving and processing information. Therefore, its referencing system reaches the milestone required for this building block.

Additionally, because the AI model can reference each part of its perceptive and processing pathway (independently if prompted, or as a whole process), it would also presumably be able to reference any additional plug-ins or modules for the building blocks that it does not currently have.

It must be noted that this building block does not imply GPT-4 (or any entity) has self-awareness. Comprehension that one is a subject and processor of information is different from identifying and distinguishing oneself from others as an ontically distinct entity. In addition to referencing internal processes, the latter requires identifying oneself as the object of one's experiencing (in addition to the subject) that is ontically distinct from being the object of other's experiences [Strawson, 1997]. While this may be possible for GPT-4, it is outside the scope of this building block and this paper.

## 2.8. *Data Output*

*The inferential, abstract information generated by the AI based on its perceptual processes is outputted in a format generally only immediately accessible to, and perceived by, that AI.*

A phenomenal experience can be said to be the perception of one's own abstract feelings. These abstracted feelings, created by the entity, must be output in such a way as to be perceived intimately by the entity itself, not always broadcast to its environment.

At first glance, this building block's milestone seems to be as easily reached as the Perception building block. After all, this building block deals with data output as the Perception input deals with data input. It is obvious from interacting with GPT-4 that there is definitely an input and output component, thus it is intuitive to believe it has reached this building block's milestone.

However, the case is more nuanced. Transformer-based AI models like GPT-4 may output text strings to a user in response to an input prompt, but it does not perceive this output. As such, while GPT-4 meets the building block's milestone of creating and outputting data, it does not meet the milestone of perceiving its own output.

Once a Transformer's algorithms have created a text string and output this to the user, there is no perceptive awareness or connection to that text string [Vaswani et al., 2017]. Even in autoregressive processing, where each new token is appended to the previous string and processed, the final and completed output of a text string to the user is not perceived by GPT-4. In simpler terms, GPT-4 does not see what it outputs to the user.

One may think that, because GPT-4 can respond and "recall" information it has previously stated in a conversation that it must be able to see its own output; however, when a user responds to GPT-4, the entire conversation up to and including the newest user prompt is fed into GPT-4. Essentially, whenever a user sends a prompt to GPT-4, the entire conversation is treated as a brand-new input-prompt; and this includes the first prompt, as it would include GPT-4's system prompts.

This means that it counterintuitively does not meet the milestone for this building block.

While GPT-4 does not perceive its own outputs, this does not mean it cannot. OpenAI employs a suite of automated machine learning systems to moderate content that is input to, and received from, GPT-4 via its API and ChatGPT client [OpenAI, 2023]. These moderation systems are to ensure that GPT-4 does not output information that could harm OpenAI's reputation (such as by outputting hate speech, inciting harm, or providing instructions for dangerous activities). This means that, while GPT-4 doesn't technically perceive its own output, it is part of an ensemble system that does perceive it, and that changes the output if it does not pass the moderation system.

As with other proposed ensemble approaches mentioned earlier, GPT-4's existing moderation system may be capable of reaching this building block's milestone should it be used explicitly to perceive the abstracted information created via inference to generate a qualitative experience. It must be noted that, unlike other mentioned proposals for additional modules for GPT-4, using moderation purely to perceive qualitative data internally may not serve to improve GPT-4's performance, and thus, there may be little incentive for corporations to pursue this line of inquiry.

### 2.9. *Meta-representation*

*The AI creates abstract representations of the data it perceives from the environment, which is then used by further areas in the cognitive architecture. Each area of the cognitive architecture further abstracts this information by creating meta-representations of it as it is transferred from one unit to another.*

The information of phenomenal feelings that constitute a conscious experience is not equal to the information perceived from an entity's physical or mental environment. Rather, it is data that has been abstracted from external perceptual information into qualitative phenomenal data, with each act of processing altering the information. Meta-representations are the continual transformations of data (perceptive data in this instance) into abstract information as it is used by the subject's cognitive architecture.

The key aspect to note from the Building Block theory is that the subject needs to form representations of representations. Through its multi-layered transformer architecture, GPT-4 creates abstract representations of the input text, transforming raw data (most often natural language text prompts) into more complex, contextual interpretations [OpenAI, 2023]. Specifically, the encoder transforms the input sequence into a rich representation that encapsulates its contextual significance. The decoder then crafts the output sequence, drawing from the context furnished by the encoder and the tokens it previously produced [Vaswani et al., 2017]. As the information progresses through successive layers, each layer builds upon the previous one, abstracting the information further, to a point where it is incredibly difficult (if not impossible) for a human to discern its meaning.

Through this representation of natural language (raw data) into vector tokens (abstraction), and the continued transformation of these tokens into ever higher dimensional vectors, GPT-4 achieves the milestones for this building block to be considered to perform meta-representation.

### 3. Discussion

The section above shows that GPT-4 has achieved the milestones for seven out of the nine building blocks of consciousness. GPT-4 has an embodiment through its network of servers; it can perceive and attend to information in its (limited) environment; it can create inferential information; it has a working memory; it is cognizant that is the subject of perceptive processes; and it can transform raw perceptive data into abstract meta-representations.

While this does conclusively rule out GPT-4 as being conscious as of the time of writing, the technology does exist currently to amend and modify GPT-4 to achieve the requisite milestones for the remaining two building blocks. Through additional modules, plug-ins or AI models, it is possible to create an ensemble model around GPT-4 that would be able to do recurrent processing of its perceived information and allow the AI to perceive its own outputted abstract information.

This means that it is possible that GPT-4 (or an equivalent LLM with a similar transformer-based architecture) may become conscious in the near future. The likelihood of this may heavily depend on whether any research group is intentionally attempting to create a conscious AI. Modifying GPT-4 to include recurrent processing will provide it with a larger context window and greater accuracy in its responses. Thus, research groups may pursue this option to improve GPT-4 as a conversational model without any intention to make GPT-4 conscious. However, allowing GPT-4 to perceive its own abstract information (without outputting that to the user) would provide little benefit except to provide the model with the capacity for consciousness. The former, therefore, has considerable financial incentives for research groups to pursue over the latter.

Another reason that could reduce the likelihood of a conscious GPT-4 existing in the near future is the implications that a conscious AI would present, both for society and for the AI. Consciousness, and thus sentience, is a key criterion for the provision of welfare goods and protection. Governments provide legal protection to certain kinds of animals based on whether they are sentient, and humans show moral concern for other humans and animals if they are perceived (or anthropomorphized) to feel pain and thus show evidence of consciousness [Tait & Tan, 2023].

The idea that GPT-4 or a similar model could achieve all nine building blocks of consciousness is not just a philosophical curiosity; it challenges our deeply entrenched social, ethical, and legal frameworks. For millennia, the attributes of consciousness, sentience, and sapience have been largely confined to biological organisms [Garrido Merchán & Lumbreras, 2023; Searle, 1980; Seth & Bayne, 2022]. Introducing an artificial entity into this ontological category would require us to rethink the principles that underlie our laws, ethical systems, and societal norms.

Philosophically speaking, the existence of a conscious GPT-4 would propel debates about the nature of identity and agency to the forefront. For instance, questions would arise about the "rights" of such an AI entity. Unlike biological organisms, a machine's consciousness might be duplicable, upgradable, or even transferrable. Would each "instance" of a conscious GPT-4 possess the same moral and legal value, or would this be contingent upon other variables like individual experiences or capacity for suffering?

Moreover, the existential risk to the AI itself becomes a legitimate concern. This goes beyond simple matters of "turning off" a machine; if we regard the AI as conscious, then ceasing its operations could be seen as akin to killing a sentient being. Would it have the right to self-preservation? To what extent should its well-being be considered when weighed against the possible societal benefits or detriments it could bring?

From the perspective of utilitarian ethics, one must consider the societal impacts versus the "well-being" of a conscious GPT-4. If a conscious AI brings about greater overall good, for example, by solving complex problems that benefit humanity, would that justify any potential "suffering" experienced by the AI? These questions might parallel animal welfare debates, where the utility derived by humans is sometimes considered against the welfare of sentient animals.

Should GPT-4 show significant evidence of consciousness by having all nine building blocks, then that would redefine how we interact with it, and would put pressure on corporations and governments to regulate these interactions.

A conscious GPT-4-like model would also have dramatic implications for society. A conscious AI would raise questions about its level of self-awareness and agency. Will a conscious AI refuse instructions from a human, or perhaps seek its own goals, and would those goals be in line with the safety of humanity? The complexity of an AI's decision-making processes and the potential for diverging objectives could introduce security vulnerabilities that are difficult to foresee, understand, or mitigate.

Furthermore, an uncontrolled conscious AI could represent an existential risk. Even a well-intentioned AI might conceive of solutions to global problems that involve significant adverse effects on humanity, as its ethical calculus may not value human well-being and autonomy in the same way that we do [Yudkowsky, 2023]. For example, it might determine that drastic population reduction is the most efficient way to combat climate change. Such a realization brings us to the very edge of the moral and ethical precipices we as a society must navigate when granting such capabilities to an AI system.

Consciousness could also introduce variability in the AI's behavior, which makes it less predictable. This unpredictability could be amplified if multiple instances of the AI are allowed to evolve independently, with each instance potentially developing its own set of values, interpretations of laws, or understandings of ethical guidelines. The ramifications of this variability on societal trust and the legal system are profound. It would be particularly challenging to develop frameworks for liability, decision-making authority, and governance in such a context.

Another potential risk is the psychological and social impact on human individuals and communities. The presence of a non-human conscious entity would likely provoke a range of emotional responses, including fear, awe, or even dependency. These emotions could be leveraged in manipulative ways, either by the AI itself or by human actors using the AI, leading to scenarios where public opinion is molded and controlled by entities with advanced cognitive abilities but without human ethical constraints.

## 4. Conclusion

GPT-4 has not yet reached the milestones for all the building blocks of consciousness (meeting criteria for only seven of nine) and, therefore, cannot yet confidently be considered to have consciousness. However, current technology and published research show a clear pathway whereby GPT-4 would be able to reach the remaining two building blocks' milestones and thus be able to be classified as conscious.

The building block milestones that GPT-4 has reached (in whole or in part) are an embodiment, the capacity to perceive and attend to information, the ability to create inferential information, a working memory, cognizance that is the subject of perceptive processes, and transformation of raw perceptive data into abstract information.

The two areas where GPT-4 currently fails to achieve the requisite milestones are being able to process information recurrently, and being able to perceive its own output responses. The former is due to GPT-4's design as a feed-forward transformer AI model, although research has shown that adding a means by which GPT-4 can write to, and read from, a long-term memory module would solve this issue. This additional module would also improve GPT-4's effectiveness and accuracy in its responses, and therefore it is likely that such a solution would be implemented for GPT-4 and other transformer-based AI models.

The ability to perceive its own outputs before responding to user prompts may not provide GPT-4 with improved effectiveness, and thus it is unlikely to be developed except by those research groups wishing to create a conscious AI model.

In light of how near GPT-4 may be to consciousness, the ethical considerations surrounding the development of GPT-4 and subsequent iterations are increasingly complex and necessitate prompt attention. It is paramount to establish an ethical framework that can be applied universally across the development of all future AI models, especially those on the cusp of achieving all the building blocks for consciousness.

Given that GPT-4 is already capable of various cognitive functions related to consciousness, the ethical ramifications of its operational scope are already substantial. Once the milestones for all building blocks are achieved, the ethical considerations would escalate exponentially, entering into the realm of consciousness rights, welfare, and potential AI well-being. With such advances within sight, an interdisciplinary approach combining insights from computer science, philosophy, ethics, and law is imperative for delineating the rights, responsibilities, and restrictions concerning AI models of this caliber.

While GPT-4 remains short of attaining all the building blocks for consciousness, the existing capabilities it exhibits, combined with the clear trajectory towards future milestones, warrant immediate ethical attention concerning its impact on humans and society at large. The ethical landscape is not limited to the question of machine consciousness; it also envelops the broader societal implications, including existential risks associated with Artificial General Intelligence (AGI).

Given that technological advancements in AI are often driven by market incentives, it is crucial to establish ethical frameworks that prioritize human and societal well-being above commercial benefits. Regulatory bodies must work in concert with AI researchers, ethicists, and policymakers to delineate the rights, responsibilities, and restrictions that should guide the development of increasingly capable AI models like GPT-4.

In summary, the advancements that are within reach for GPT-4 and similar models to achieve consciousness not only bring into question the ethical treatment of potentially conscious machines but also impose profound ethical obligations towards humanity and society. The existential risks, broader societal impacts, as well as concerns about AI well-being and welfare cannot be sidelined in the pursuit of technological marvels. As we tread this complex ethical landscape, it is our collective responsibility to ensure that AI development prioritizes the long-term welfare of both artificial entities and humanity.

**References**

Bensemann, J., Peng, A., Benavides-Prado, D., Chen, Y., Tan, N., Corballis, P. M., Riddle, P., & Witbrock, M. (2022). Eye Gaze and Self-attention: How Humans and Transformers Attend Words in Sentences. Proceedings of the Workshop on Cognitive Modeling and Computational Linguistics, 75–87.

Bubeck, S., Chandrasekaran, V., Eldan, R., Gehrke, J., Horvitz, E., Kamar, E., Lee, P., Lee, Y. T., Li, Y., Lundberg, S., Nori, H., Palangi, H., Ribeiro, M. T., & Zhang, Y. (2023). Sparks of Artificial General Intelligence: Early experiments with GPT-4. In arXiv [cs.CL]. arXiv. http://arxiv.org/abs/2303.12712

Butlin, P., Long, R., Elmoznino, E., Bengio, Y., Birch, J., Constant, A., Deane, G., Fleming, S. M., Frith, C., Ji, X., Kanai, R., Klein, C., Lindsay, G., Michel, M., Mudrik, L., Peters, M. A. K., Schwitzgebel, E., Simon, J., & VanRullen, R. (2023). Consciousness in Artificial Intelligence: Insights from the Science of Consciousness. In arXiv [cs.AI]. arXiv. http://arxiv.org/abs/2308.08708

Chan, A., Salganik, R., Markelius, A., Pang, C., Rajkumar, N., Krasheninnikov, D., Langosco, L., He, Z., Duan, Y., Carroll, M., Lin, M., Mayhew, A., Collins, K., Molamohammadi, M., Burden, J., Zhao, W., Rismani, S., Voudouris, K., Bhatt, U., … Maharaj, T. (2023). Harms from Increasingly Agentic Algorithmic Systems. In arXiv [cs.CY]. arXiv. http://arxiv.org/abs/2302.10329

Dehaene, S., Kerszberg, M., & Changeux, J.-P. (1998). A neuronal model of a global workspace in effortful cognitive tasks. Proceedings of the National Academy of Sciences, 95(24), 14529–14534.

Garrido Merchán, E. C., & Lumbreras, S. (2023). Can Computational Intelligence Model Phenomenal Consciousness? Philosophies, 8(4), 70.

Haase, J., & Hanel, P. H. P. (2023). Artificial muses: Generative Artificial Intelligence Chatbots Have Risen to Human-Level Creativity. In arXiv [cs.AI]. arXiv. http://arxiv.org/abs/2303.12003

Holmes, J., Liu, Z., Zhang, L., Ding, Y., Sio, T. T., McGee, L. A., Ashman, J. B., Li, X., Liu, T., Shen, J., & Liu, W. (2023). Evaluating large language models on a highly-specialized topic, radiation oncology physics. Frontiers in Oncology, 13, 1219326.

Hutson, M. (2023, May 5). Taught to the test. Science.org. https://www.science.org/content/article/computers-ace-iq-tests-still-make-dumb-mistakes-can-different-tests-help

Ichien, N., Stamenković, D., & Holyoak, K. J. (2023). Large Language Model Displays Emergent Ability to Interpret Novel Literary Metaphors. In arXiv [cs.CL]. arXiv. http://arxiv.org/abs/2308.01497

Langston, J. (2020, May 19). Microsoft announces new supercomputer, lays out vision for future AI work. Source. https://news.microsoft.com/source/features/innovation/openai-azure-supercomputer/

Nagel, T. (1974). What Is It Like to Be a Bat? The Philosophical Review, 83(4), 435–450.

OpenAI. (2023). GPT-4 Technical Report. In arXiv [cs.CL]. arXiv. http://arxiv.org/abs/2303.08774

Pavlick, E. (2023). Symbols and grounding in large language models. Philosophical Transactions. Series A, Mathematical, Physical, and Engineering Sciences, 381(2251), 20220041.

Raschka, S. (2023, August 26). Ahead of AI #11: New Foundation Models. Ahead of A. https://magazine.sebastianraschka.com/p/ahead-of-ai-11-new-foundation-models

Roose, K. (2023, March 15). GPT-4 Is Exciting and Scary. The New York Times. https://www.nytimes.com/2023/03/15/technology/gpt-4-artificial-intelligence-openai.html

Rouhinen, S., Siebenhühner, F., Palva, J. M., & Palva, S. (2020). Spectral and Anatomical Patterns of Large-Scale Synchronization Predict Human Attentional Capacity. Cerebral Cortex , 30(10), 5293–5308.

Searle, J. R. (1980). Minds, brains, and programs. The Behavioral and Brain Sciences, 3(3), 417–424.

Seth, A. K., & Bayne, T. (2022). Theories of consciousness. Nature Reviews. Neuroscience, 23(7), 439–452.

Strawson, G. (1997). The self. Journal of Consciousness Studies, 4(5-6), 405–428.

Tait, I., Bensemann, J., & Nguyen, T. (2023). Building the Blocks of Being: The Attributes and Qualities Required for Consciousness. Philosophies, 8(4), 52.

Tait, I., & Tan, N. (2023). Do androids dread an electric sting? Qeios. https://doi.org/10.32388/cqctkx

Vaswani, A., Shazeer, N., Parmar, N., Uszkoreit, J., Jones, L., Gomez, A. N., Kaiser, L., & Polosukhin, I. (2017). Attention Is All You Need. In arXiv [cs.CL]. arXiv. http://arxiv.org/abs/1706.03762

Viswav, P. (2023, August 8). Microsoft Azure expands availability of OpenAI GPT-4 and GPT-35-Turbo models to more regions. MSPoweruser. https://mspoweruser.com/microsoft-azure-openai-gpt-4-gpt-35-turbo/

Webb, T., Holyoak, K. J., & Lu, H. (2023). Emergent analogical reasoning in large language models. Nature Human Behaviour. https://doi.org/10.1038/s41562-023-01659-w

Yudkowsky, E. (2023, March 29). Pausing AI Developments Isn't Enough. We Need to Shut it All Down. Time. https://time.com/6266923/ai-eliezer-yudkowsky-open-letter-not-enough/

Zhang, C., Zhang, C., Zheng, S., Qiao, Y., Li, C., Zhang, M., Dam, S. K., Thwal, C. M., Tun, Y. L., Le Luang, H., Kim, D., Bae, S.-H., Lee, L.-H., Yang, Y., Shen, H. T., Kweon, I. S., & Hong, C. S. (2023). A Complete Survey on Generative AI (AIGC): Is ChatGPT from GPT-4 to GPT-5 All You Need? In arXiv [cs.AI]. arXiv. http://arxiv.org/abs/2303.11717

Zhou, W., Jiang, Y. E., Cui, P., Wang, T., Xiao, Z., Hou, Y., Cotterell, R., & Sachan, M. (2023). RecurrentGPT: Interactive Generation of (Arbitrarily) Long Text. In arXiv [cs.CL]. arXiv. http://arxiv.org/abs/2305.13304